\documentclass[conference]{IEEEtran}
\IEEEoverridecommandlockouts
\usepackage{cite}
\usepackage{amsmath,amssymb,amsfonts}
\usepackage{graphicx}
\usepackage{verbatim}
\usepackage{textcomp}
\usepackage{xcolor}
\usepackage[a4paper, total={184mm,239mm}]{geometry}
\def\BibTeX{{\rm B\kern-.05em{\sc i\kern-.025em b}\kern-.08em
    T\kern-.1667em\lower.7ex\hbox{E}\kern-.125emX}}

\usepackage{subcaption}
\usepackage{flushend}
\usepackage{multirow}
\usepackage{tabularx}
\usepackage{dsfont}
\usepackage{mathtools}
\usepackage{bm}

\usepackage{algorithm}
\usepackage[noend]{algpseudocode}
\usepackage{xcolor}
\usepackage{siunitx}
\usepackage{array}
\newcolumntype{P}[1]{>{\centering\arraybackslash}p{#1}}

\title{Variability-Aware Training and Self-Tuning of Highly Quantized DNNs for Analog PIM\\
}

\author{\IEEEauthorblockN{1\textsuperscript{st} Zihao Deng}
\IEEEauthorblockA{\textit{Department of Electrical and Computer Engineering} \\
\textit{University of Texas at Austin}\\
Austin TX, USA \\
zihaodeng@utexas.edu}
\and
\IEEEauthorblockN{2\textsuperscript{nd} Michael Orshansky}
\IEEEauthorblockA{\textit{Department of Electrical and Computer Engineering} \\
\textit{University of Texas at Austin}\\
Austin TX, USA \\
orshansky@utexas.edu}
}

\begin{document}
\maketitle


\begin{abstract}
   DNNs deployed on analog processing in memory (PIM) architectures are subject to fabrication-time variability.    
   We developed a new joint variability- and quantization-aware DNN training algorithm for highly quantized analog PIM-based models that is significantly more effective than prior work. 
      It outperforms variability-oblivious and post-training quantized models on multiple computer vision datasets/models. 
      For low-bitwidth models and high variation, the gain in accuracy is up to $35.7\%$ for ResNet-18 over the best alternative. 
  
        We demonstrate that, under a realistic pattern of within- and between-chip components of variability, training alone is unable to prevent large DNN accuracy loss (of up to 54\% on CIFAR-100/ResNet-18). 
  We introduce a self-tuning DNN architecture that dynamically adjusts layer-wise activations during inference and is effective in reducing accuracy loss to below 10\%.
\end{abstract}

\section{Introduction}
Analog hardware accelerators for DNNs, based on the processing in memory (PIM) architectures, 
perform computation within a memory array that dramatically reduces data movement, eliminates the von Neumann bottleneck, and the associated costs in energy and latency  \cite{cosemans2019towards}.
A key compute kernel needed for DNN inference is a matrix-vector multiplication (MVM) array. 
Analog PIM accelerators rely on the possibility of efficient dot-product calculation using analog-based current summation and have been shown for multiple device technologies, including Resistive RAM, Flash, and MRAM \cite{Long2019DesignOR,Charan2020AccurateIW,Kang2021SFLASHAN,Tao2019CINTA,Wang2020ANM,Patil2019AnMD}.
DNN weights are represented by analog conductances (resistances) of memory cells. With the wordline voltages representing activations, the output current in a bitline is proportional to the activation-weight dot product \cite{Chakraborty2020ResistiveCA}.

Though, in theory, a continuous range of conductance is possible, in practice, there is a discrete set of feasible conductance values.
The number of feasible conductance levels depends on technology. 
RRAM and Flash allow more than two states per cell. 
For example, there is currently a production-ready 5-bits per cell Flash technology available {\cite{PLCFlash}}. 
At the same time, the precision of activations is limited by the resolution of DACs (producing wordline voltages/input activations) and ADCs (quantizing the dot-product results).     

Conventional DNN training is based on single-precision representation of weights and activations. 
Aggressive quantization of both activations and weights means that PIM-based DNNs 
need to rely on quantization-aware training (QAT) that directly incorporates quantization error into training \cite{Krishnamoorthi2018QuantizingDC,Rastegari2016XNORNetIC,Li2016TernaryWN,Courbariaux2015BinaryConnectTD,Bulat2019XNORNetIB}.
As we demonstrate, deploying models on highly quantized platforms without QAT, or using naive post-training quantization, leads to drastic quality loss.  

Yet currently available QAT algorithms focus on issues around digital quantizaiton. They ignore another significant challenge for practical adoption of analog PIM DNN accelerators: how to handle the non-ideality of weight representations as physical variables, e.g., conductances. The use of bit-cell conductance as a non-binary quantity intrinsically makes it subject to being impacted by the randomness of semiconductor fabrication. Nanometer scale semiconductor devices exhibit large variability across multiple spatial scales \cite{stine1997analysis}. As a result, the programmed conductances of memory cells will deviate from their intended values. 
These deviations in memory cells' conductances are then translated into variations of weights inside DNNs once they are deployed on PIM devices. 
Such parameter variation significantly degrades the performance of DNNs. 

\emph{The main focus of the paper is the development of improved variability-aware training (VAT) methods and DNN network-structure innovation that mitigate such degradation, specifically, in the context of highly-quantized deployments.} 
We describe a training approach for the joint quantization-aware and variability-aware training (QAVAT).
{Training quantized neural networks involves finding the derivative of a piecewise constant (threshold) function whose gradient vanishes almost everywhere.}
Similar to prior QAT works \cite{Courbariaux2015BinaryConnectTD,Li2016TernaryWN,Rastegari2016XNORNetIC,Bulat2019XNORNetIB}, QAVAT uses the straight-through estimator to back-propagate gradients through {the} hard threshold function as though it were an identity function.
To train robust networks against physical variability, the practical strategies explored by most researchers use implicit robustification.
These techniques inject noise during the forward pass of the backpropagation algorithm to simulate the effect of parameter variation \cite{Long2019DesignOR,Charan2020AccurateIW,Zhou2020NoisyMU}. 
In every forward propagation, a noise vector is sampled from a pre-defined distribution and applied to the parameters to calculate loss. 

Our training algorithm also relies on implicit robustification through injection of variability during training. However, we advance the state-of-the-art in several aspects. 
Prior work has not addressed the question of an optimal sampling strategy. We describe a sampling mechanism ensuring that the gradient estimates are not biased and have reduced variance, using the techniques of reparameterization and multi-sampling. 

\emph{Further, we point out that earlier work in VAT has not made a critical distinction in its treatment of variability structure.}
It assumed independence of variability, failing to capture its spatial structure. 
In real manufacturing, chip-to-chip variation is typically responsible for a sizable contribution to overall variability. 
We show that in the presence of spatially correlated variability, training-time only solutions fail to prevent large DNN accuracy loss. 
To effectively deal with correlated variability, we introduce a \emph{self-tuning DNN architecture} realized via the insertion of light-weight self-tuning modules into DNN models. 
Our self-tuning DNN architecture is general and different from circuit-specific solutions \cite{Joshi2020AccurateDN,Gallo2018CompressedSW}.

In summary, the contributions of this work are:
\begin{itemize}
    \item A novel implementation of joint quantization- and variability-aware DNN training (QAVAT). It significantly outperforms variability-oblivious QAT and post-training quantized VAT (PTQ-VAT). For ternary-weight ResNet-18 on CIFAR-100, on average, QAVAT achieves $42\%$ and $14\%$ higher accuracy than PTQ-VAT and QAT. 
    \item For realistic variability with spatial correlations, training is unable to prevent large accuracy loss, e.g. of $50\%$ for ResNet-18/CIFAR-100. 
    \item A general self-tuning architecture, suitable for an arbitary DNN, that dynamically adjusts its layer outputs during inference and is very effective in reducing accuracy loss under correlated variability (from $50\%$ to below $10\%$ for ResNet-18/CIFAR-100).
    \item A design space exploration of self-tuning including its size-quality trade-off.
\end{itemize}


\section{Quantization- and Variation-Aware Training: Algorithm and Models}

\subsection{Quantization- and Variation-Aware Gradient Estimation}
Our algorithm utilizes Monte Carlo-style injection of uncertainty during network training as a way of making the resulting network robust against physical variations. 
The Straight-Through Estimator (STE) is used jointly with uncertainty sampling to estimate the descending direction.
As a way to formalize key aspects of such training, we introduce a variational computational graph to describe the computations of QAVAT, Fig. \ref{fig:CompGraphQAVAT}. 
The main innovative component of the graph is its specification of the variability sampling process. 

The key aspect of the sampling-based gradient-descent algorithm is the injection of variability. 
It is important to ensure that the gradients computed for weight updates are unbiased with respect to variability sampling. 
Formally, we seek a graph such that the gradient ($g$) computed on each batch of training samples satisfies
$\mathbb{E}(g) = \nabla_{w}\mathbb{E}_{x,y,\delta_{w}}[L(M_{w+\delta_w}(x),y)]$, where 
{$M_w(\cdot)$ represents an arbitrary DNN model parameterized by $w$, $L$ is the loss function, and}
the distribution of loss depends on data $x$, labels $y$, and weight variation $\delta_w$.
{Note that the distribution of $\delta_w$ usually depends on $w$.}
\begin{center}
    \centering
    \captionsetup{type=figure}
    \includegraphics[width=.48\textwidth]{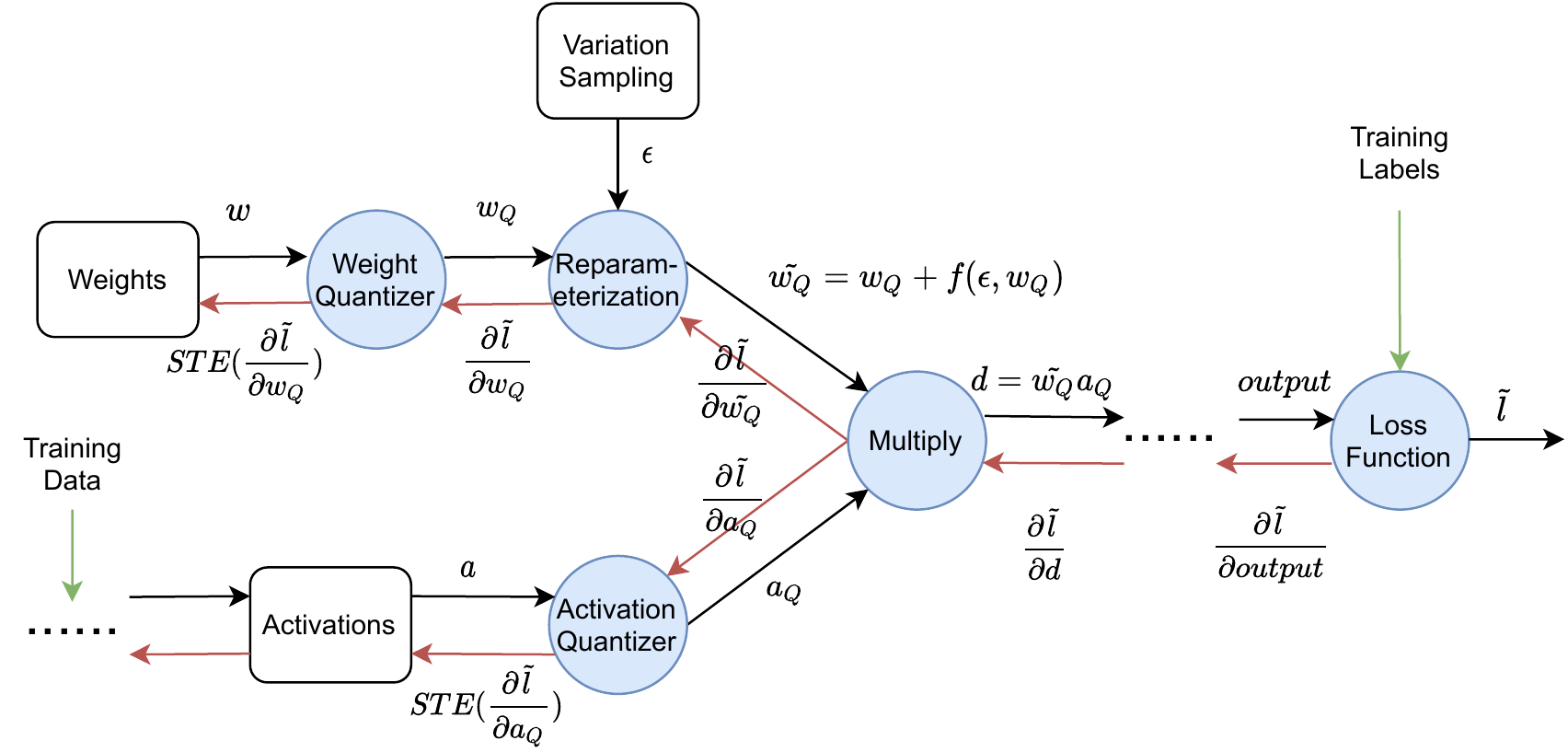}
    \caption{Computational graph of joint quantization- and variation-aware training. Key elements: reparameterization and multi-variation Monte Carlo sampling for variability, STE for quantization. Multi-variation sampling is done by accumulating the gradients of multiple forward (black arrows) and backward (red arrows) passes through the computational graph.}
	\label{fig:CompGraphQAVAT}
\end{center}%

\emph{The choice of a sampling distribution for variability is critical for computing an unbiased estimate of the gradient.} 
When using an auto-differentiation system, the simplest approach is to sample variations $\delta_{w}$ numerically and add them onto weights.
Unfortunately, this procedure computes a biased estimator of the true gradients. 
In Eq. \ref{eq:bias}, the expected value of the gradient on the left side is what the procedure yields in backpropagation.
However, it does not capture the impact of $w$ on the distribution of $\delta_w$.
The true gradient is on the right side.
\begin{equation}
\begin{aligned}
    \mathbb{E}_{x,y,\delta_{w}}[\nabla_{w}L(M_{w+\delta_w}(x),y)] \neq \nabla_{w}\mathbb{E}_{x,y,\delta_{w}}[L(M_{w+\delta_w}(x),y)]
    \label{eq:bias}
\end{aligned}
\end{equation}
To overcome this problem, we reformulate the computational graph using reparameterization \cite{Kingma2014AutoEncodingVB}, which allows computing an unbiased estimator of true gradient\footnote{We have seen no other work on variability-aware training that has described the need for reparameterization.}.
Reparameterization assumes that there exists 
a differentiable function $f$ such that
$f(w,\epsilon)$ generates the same distribution as $\delta_w$ with a new random variable $\epsilon$ that is independent of $w$. 
As an example, $f(\epsilon,w)=\epsilon w$, with $\epsilon\sim N(0,\sigma^2)$, is a valid reparameterization of $\delta_w \sim N(0,\sigma^2 |w|^2)$.
With reparameterization, we have:
\begin{equation}
\begin{aligned}
\nabla_{w}\mathbb{E}_{x,y,\delta_{w}}[L(M_{w+\delta_w}(x),y)] \\
= \nabla_{w}\mathbb{E}_{x,y,\epsilon}[L(M_{w+f(w,\epsilon)}(x),y)]\\
= \mathbb{E}_{x,y,\epsilon}[\nabla_{w}L(M_{w+f(w,\epsilon)}(x),y)]
\label{eq:unbiased}
\end{aligned}
\end{equation}
Unlike Eq. \ref{eq:bias}, in the last step of Eq. \ref{eq:unbiased} the gradient operator can be safely moved inside the expectation because the reparameterized variability $\epsilon$ is now independent of $w$.
Therefore, the back-propagated gradient $\nabla_{w}L(M_{w+f(w,\epsilon)}(x),y)$ is unbiased. 

It is also important to compute a low-variance estimator of the gradient. 
The variance of the estimator is partly determined by how many $\epsilon$s in Fig. \ref{fig:CompGraphQAVAT} are sampled per optimization step. 
Without discussing the importance of using a low-variance estimator, prior work uses a single sample per update \cite{Long2019DesignOR,Charan2020AccurateIW,Zhou2020NoisyMU}. 
As we show in experiments, a measurable benefit can be achieved when multiple samples are used.
We refer to this as multi-sampling.

The injection of variability, as shown in Fig. \ref{fig:CompGraphQAVAT}, is applied after the quantization of weights.
We now describe our quantization scheme and introduce the use of STE and show how gradients are estimated when variability is present.
We use the uniform symmetric quantizer \cite{Dai2021VSQuantPS} for both activations and weights. 
Let $x$ represent the tensor to be quantized (weight or activation), $k$ be the bitwidth, and $\Delta$ be the scaling factor. Let $x_{Q}$ be the quantized tensor and $x_{D}$ be its dequantized value. 
The quantization strategy is summarized in Eq. \ref{eq:quantization}.
\begin{equation}
\begin{aligned}
x_{Q} &= clip(\lfloor\frac{x}{\Delta}\rceil,-2^{k-1}+1,2^{k-1}-1)\\
x_{D} &= x_{Q}\cdot\Delta \\
\tilde{x_{D}} &= x_{D} + f(\epsilon,x_{D})
\label{eq:quantization}
\end{aligned}
\end{equation}
In the above equation, when $x$ stands for weights, $f$ and $\epsilon$ are its variability reparameterization.
$\tilde{x_{D}}$ represents the variability-impacted weights after dequantization. 
When $x$ stands for activations, $f$ is a dummy function that always returns zero because no explicit
variability is assumed on activations.
$\tilde{x_{D}}$ in this case is equal to $x_{D}$.

We use the minimum mean squared error (MMSE) heuristic\cite{Choukroun2019LowbitQO} to compute the scaling factors of weights.  
By default, we compute them only at the beginning of training.
We found that more frequent updates only improve results marginally.
We also use static method for activation quantization in which the scaling factors are fixed during training.
Dynamic methods \cite{Krishnamoorthi2018QuantizingDC,Choi2018PACTPC}, without extensive fine-tuning (e.g., without adding regularization of clipping thresholds in \cite{Choi2018PACTPC}), have shown degraded performance compared to a static estimation scheme.


 
To compute the gradients of quantized tensors under variability, STE is used:
\begin{equation}
\begin{aligned}
\frac{\partial L}{\partial x} &= \frac{\partial L}{\partial \tilde{x_{D}}} \frac{\partial \tilde{x_{D}}}{\partial x_D}\frac{\partial x_D}{\partial x} \overset{\text{STE}}{\approx} \frac{\partial L}{\partial \tilde{x_{D}}} (1+\frac{\partial f(\epsilon,x_{D})}{\partial x_D})
\label{eq:STE}
\end{aligned}
\end{equation}
$L$ denotes the training loss and $x$ denotes weights and activations. 
The entire QAVAT is described in Algorithm \ref{algo:QAVAT}.

\begin{algorithm}
\caption{Multi-Variation Sampling QAVAT}\label{algo:QAVAT}
\begin{algorithmic}[1]
\item \textbf{Given}
\item \hspace{20pt} $n$: number of variation samples
\item \hspace{20pt} $w_t$: network parameters at step $t$
\item \hspace{20pt} $T$: number of iterations, $\eta$: step size
\item \hspace{20pt} $L(w,B)$: loss of network $w$ on mini-batch $B$ 
\item \hspace{20pt} $\nabla$: gradient estimation using STE
\Procedure{}{}
\For {$t$ in $0:T-1$}
    \State Sample a mini-batch of training data $B$
    \State $l\gets0$
    \For {$i$ in $1:n$}
        \State Sample $\epsilon_i$ and $l \gets l + L(f( w_t,\epsilon_i),B)$
    \EndFor
    \State ${w}_{t+1}\gets w_{t} -\eta\nabla_{w_{t}}l$
\EndFor
\State \textbf{return } $ w_T$
\EndProcedure
\end{algorithmic}
\end{algorithm}

\subsection{Joint Within- and Between-Chip Variability Modelling}
\emph{Prior work in variability-aware training has not addressed an important issue: realistic patterns of variability found in semiconductor manufacturing have pronounced and significant structure.}
Instead, earlier work assumes that variability is generated by independent zero-mean Gaussian sources \cite{Long2019DesignOR,Charan2020AccurateIW,Zhou2020NoisyMU}. 
These models account for only one of the prevalent types of variation, commonly referred to as within-chip variation. 
Within-chip variation, which we denote by 
$\epsilon_{W}$, 
captures the spatially uncorrelated stochastic nature of variability patterns that impact individual memory cells on a given chip. 

In reality, there is a significant component of variation that occurs on a chip-to-chip basis (referred to as the between-chip variation). It captures the variability observed between the mean values of all cell characteristics per chip. It is typically responsible for a sizable contribution to the overall variability.
The two patterns of variation can be modeled using an additive model. 
We introduce a  fully-correlated Gaussian variable $\epsilon_{B}$ to represent the between-chip variation.
In the realistic case, where both types of variations exist, the additive effect of $\epsilon_{W}$ and $\epsilon_{B}$ leads to partially correlated variations of different network parameters.
We use $\sigma_{W}$ and $\sigma_{B}$ to denote the standard deviation of $\epsilon_W$ and $\epsilon_B$, respectively.
The total variance of variations is $\sigma_{tot}^2=\sigma_{W}^2+\sigma_{B}^2$.

The sampling of variations is done in their reparameterized variable space.
To sample a single variation ${\delta_w} \in \mathds{R}^d$ on weights ${w} \in \mathds{R}^d$, we construct its reparameterization variable $\epsilon$ in the following way.
First, a random variable $\epsilon_{B} \sim N(0,\sigma_{B}^2)$ is sampled to capture the between-chip variability. 
Then, $d$ iid random variables $\epsilon_{W,i} \sim N(0,\sigma_{W}^2)$ are sampled to capture the within-chip variability.
The $i$-th entry of ${\delta_w}$ is calculated as ${\delta_{w,i}}=f(\epsilon_i,{w}_i)$, where $\epsilon_i=\epsilon_{B}+\epsilon_{W,i}$. $f$ is a reparameterization of ${\delta_w}$.


Following prior work, we model weight variation $\delta_w$ as $\delta_w \sim N(0,\sigma(w))$.
Further, we explore two models of $\sigma(w)$ \cite{Long2019DesignOR,Joshi2020AccurateDN}, both being of practical interest.
In the first model \cite{Long2019DesignOR}, the standard deviation is proportional to the magnitude of a weight: $\sigma(w)=\sigma |w|$. 
We refer to it as the ``weight-proportional variance'' model.
In the second model \cite{Joshi2020AccurateDN}, all variations in the same layer ($l$-th) share a fixed standard deviation $\sigma^l$, which only depends on the largest weight $w_{max}^l$ in that layer: $\sigma(w)=\sigma |w_{max}^l|$. 
We refer to it as the ``layer-fixed variance'' model.
We use reparameterization $f(\epsilon,w)=\epsilon w$ for the former, and $f(\epsilon,w)=\epsilon w_{max}^l$ for the latter.
\section{Self-Tuning DNNs}
\label{sec:ST}
\subsection{Better Training is Not Enough against Realistic Variation Patterns}
In the presence of significant between-chip variation, training-time methods of producing a robust DNN model appear to fail, even when they succeed at rectifying the detrimental effect of within-chip variation (shown later in our experiments). 

It is clear that within- and between-chip variations impact the network accuracy differently. We can qualitatively understand the difference in behavior using the following model. 
For any given chip, between-chip variation shifts all weights and, by extension, all dot-products and all activations, in the same positive/negative direction.
This is in contrast to the impact of within-chip variation where, on any given chip, some weights are shifted positively while others negatively, effectively creating an averaging effect in the computed dot-products/activations.

\subsection{Self-Tuning DNNs}
We have developed a novel general method for mitigating the impact of the between-chip variation\footnote{Though we limit our formal treatment to fabrication-time between-chip variation, the proposed self-tuning architecture can be generalized to compensate for any correlated weight variation, e.g., due to temperature drifts or aging.}
that achieves robustness via insertion of small modules in networks.
\emph{The proposed self-tuning (ST) architecture is suitable for an arbitary DNN and dynamically adjusts its layer outputs during inference.}

The architecture estimates layer-wise activation deviations and corrects for them.
Suppose the analog PIM array implements a MVM of inputs ${x}\in\mathbb{R}^{d_{in}}$ and weight matrix ${W}\in\mathbb{R}^{d_{out}\times d_{in}}$ to produce ${y}\in\mathbb{R}^{d_{out}}$. 
Let $W_{max}$ be the largest value in $W$.
We initially let $\sigma_W=0$ and show that the structure of self-tuning modules depends on the form of variability.
\begin{figure}[h]
     \centering
         \centering
         \includegraphics[width=.48\textwidth]{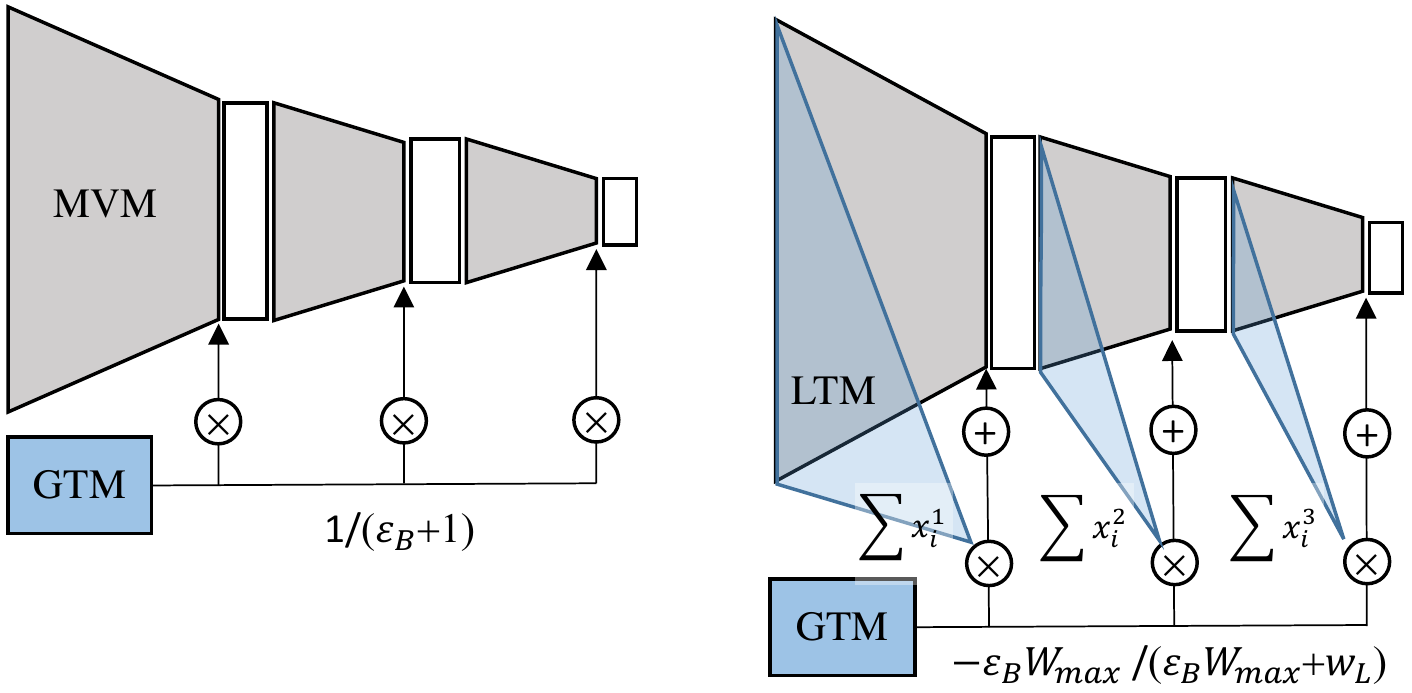}
         \caption{Self-tuning architectures for variability with weight-proportional variance (left) and layer-fixed variance (right).}
         \label{fig:selfCalib}
\end{figure}

\begin{figure}[h]
     \centering
         \centering
         \includegraphics[width=.48\textwidth]{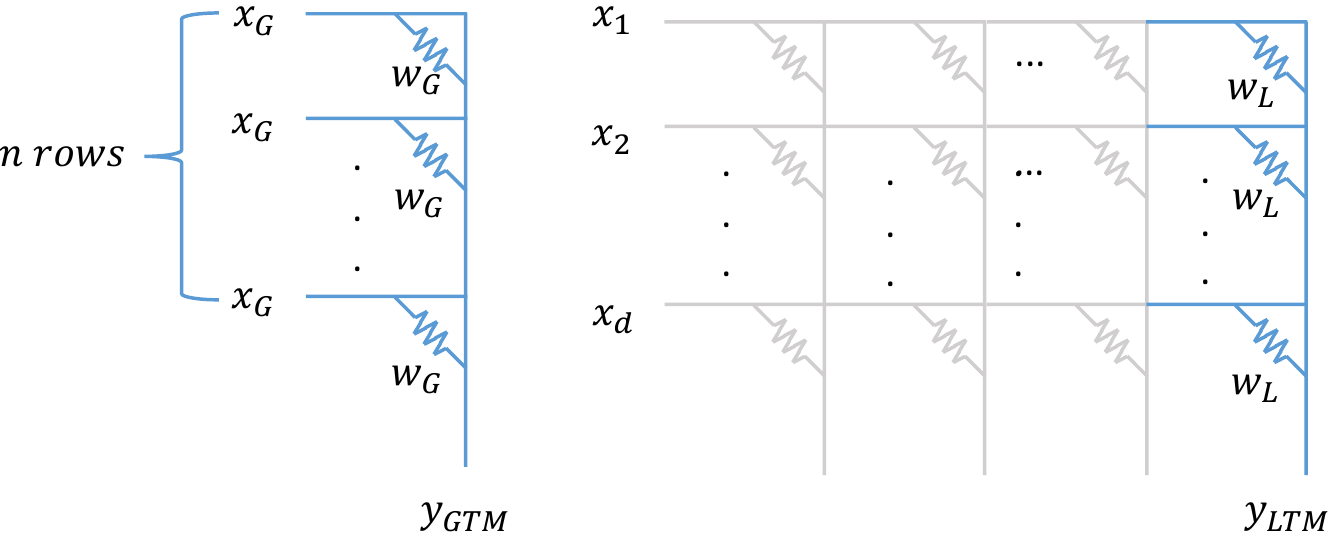}
         \caption{Circuit implementation of GTM (left) and LTM (right). GTM estimates the model-wise between-chip variation and LTM estimate the layer-wise input activation sums.}
         \label{fig:CM_LCM}
\end{figure}

If the variation follows the weight-proportional variance model, the noisy components of output are $\tilde{y}_i= \sum_{j=1}^{d_{in}} (W_{i,j}+\epsilon_{B} W_{i,j})  x_{j} = (1+ \epsilon_{B}) y_i$.
Thus, the correction term that needs to be applied to every component of $y$ is the same: $1/(1 + \epsilon_{B})$.
We introduce a Global Tuning Module (GTM) to estimate $\epsilon_{B}$.
As shown in Fig. \ref{fig:CM_LCM}, the GTM is a single column of the analog PIM array with $n$ cells.
The fixed inputs $x_G$ (voltages) and weights $w_G$ (conductances) are selected based on circuit implementation.
The variation-free value of the GTM output is $y_0=nw_Gx_G$ and is stored digitally.
The true output of GTM module, under variation, is $y_{GTM}=y_0(1+\epsilon_{B})$.
By measuring $y_{GTM}$ and dividing it by $y_0$ in the digital domain, we get the exact value of $\epsilon_{B}$.
When within-chip variability is present ($\sigma_w\neq 0$), GTM is an unbiased estimator of $\epsilon_{B}$.

The self-tuning of variations with layer-fixed variance is more costly. In this case, estimating $\epsilon_{B}$ is insufficient to recover each layer's outputs as their error also depends on input activations.
Under the layer-fixed variance model, the noisy output is $\tilde{y}=({W}+\epsilon_{B}W_{max}){x}$. 
The error in each output dimension is
$(\tilde{y}_i-y_i)=\epsilon_{B}W_{max}\sum_{j=1}^{d_{in}}x_{j}$.
$\epsilon_B$ can be estimated using GTM.
$W_{max}$ is stored digitally.
The remaining term $\sum_{j=1}^{d_{in}}x_{j}$ is estimated on a per-layer basis.

To estimate the sum of activations, we introduce a new Layer Tuning Module (LTM) that is added to every layer of the model, Fig. \ref{fig:selfCalib} (right).
The circuit implementation of LTM is a single column of the PIM memory array with weights (conductances) of all cells set to the same value $w_L$, Fig. \ref{fig:CM_LCM} (right). 
$w_L$ is stored digitally.
Under variability, the output of LTM is $y_{LTM} = (w_L+\epsilon_{B}W_{max})\sum_{j=1}^{j=d_{in}}x_j$. 
The tuning is completed in the digital domain where the correction factor $({\epsilon}_{B}W_{max}/(w_L+{\epsilon}_{B}W_{max}) \cdot y_{LTM})$ is subtracted from every component of $\tilde{y}$. 
{Allocating more LCM columns further reduces estimation variance and improves results. We use LCM=$n$ to denote the use of LCMs with $n$ columns.}

The area overhead of self-tuning architectures is small.
LTMs are instantiated per MVM array, adding columns to each array.
Assuming a single array of size $512\times512$ \cite{Chakraborty2020ResistiveCA}, the overhead is $0.2\%$ if LCM=1, and is $3.1\%$ if LCM=16.
The overhead from GTM is negligible because only one GTM is needed per chip. The largest number of GTM cells we experiment with is $10^5$. This is less than $0.1\%$ of reported  demonstrations of analog PIM architectures in hardware \cite{Chakraborty2020ResistiveCA}.
When deployed in a self-tuning mode, the network, without the self-tuning modules, is first trained using QAVAT, with Monte Carlo sampling capturing only the within-chip variation. The self-tuning modules are then appended to the trained model.

\section{Experiments}
We study the effectiveness of QAVAT on several widely-used models and testsuits: LeNet-5 on MNIST, VGG-11 on CIFAR-10, and ResNet-18 on CIFAR-100. 
We compare QAVAT with post-training quantized VAT (PTQ-VAT) models and variability-oblivious QAT models.
These models are trained under 5 different values of variability standard deviation: $\sigma_{W}\in\{0.1,0.2,0.3,0.4,0.5\}$.
The notation ``AxWy'' represents a model with x-bit activations and y-bit weights.
We explore a number of low-bitwidth combinations: A2W2 for LeNet-5, A4W2 and A8W4 for VGG-11 and ResNet-18.
By default, we use a single-sample version of QAVAT.
The resulting quality (robustness) of each trained model under variability is evaluated using a testing run involving 2000 samples of the variability vectors. The mean test accuracy of the resulting 2000 models is reported.
For consistency, the scaling factors of activations in PTQ-VAT are calibrated by the moving average of min-max values on batches of training data \cite{Krishnamoorthi2018QuantizingDC}. 
The scaling factors of weights are computed using MMSE\cite{Choukroun2019LowbitQO}.

Experiments are conducted under two scenarios of variability structure. The first scenario represents the case when between-chip variability is negligible and only within-chip variation is present. The second scenario studies a more realistic setting in which both between-chip and within-chip variations exist.

\subsection{Scenario 1: Within-chip Variation is Dominant}
\begin{figure}[ht!]
     \centering
     \begin{subfigure}[b]{0.24\textwidth}
         \centering
         \includegraphics[width=\textwidth]{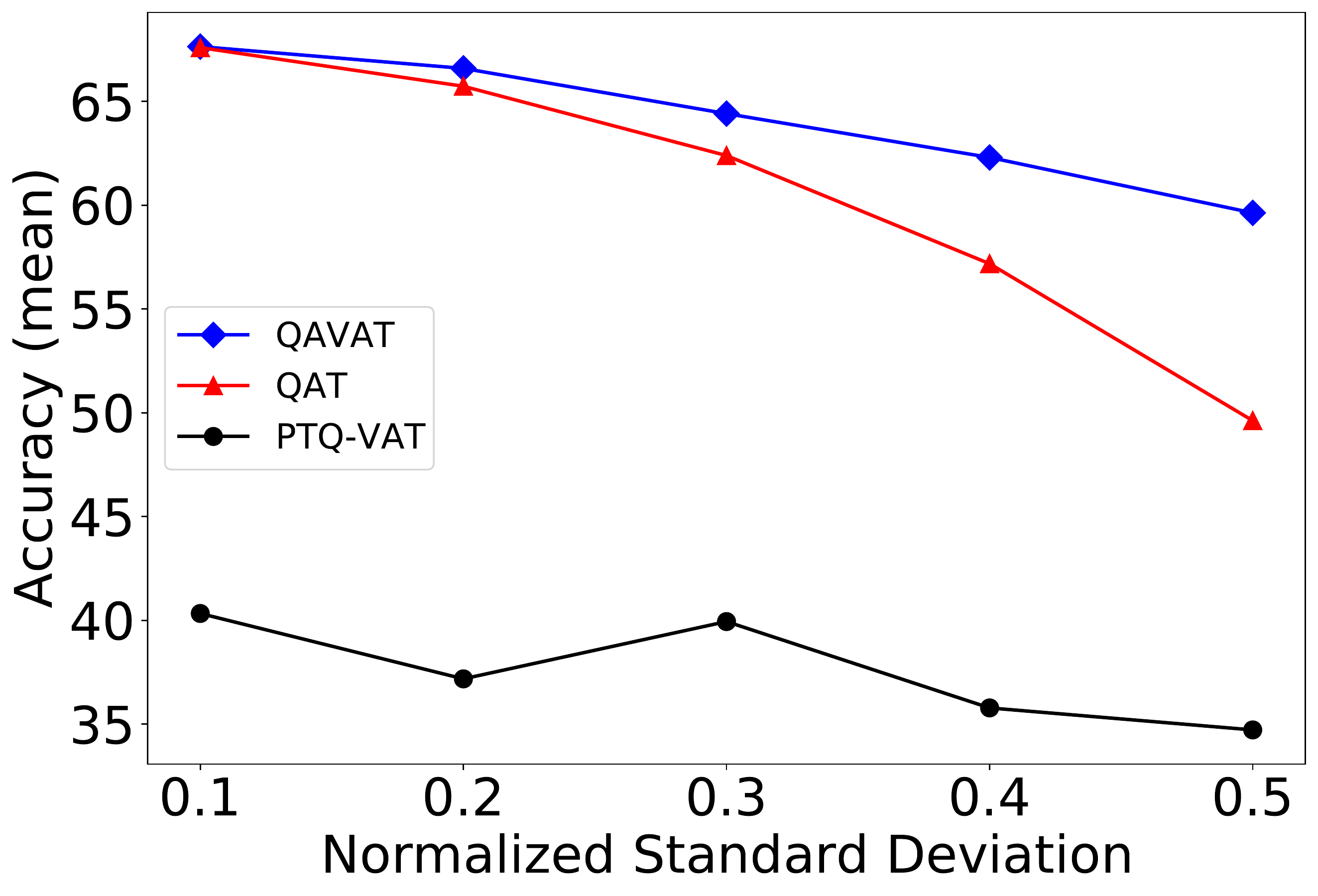}
         \caption{A4W2, weight-proportional}
         \label{fig:resnet_A4W2_mul}
     \end{subfigure}
     \hfill
     \begin{subfigure}[b]{0.24\textwidth}
         \centering
         \includegraphics[width=\textwidth]{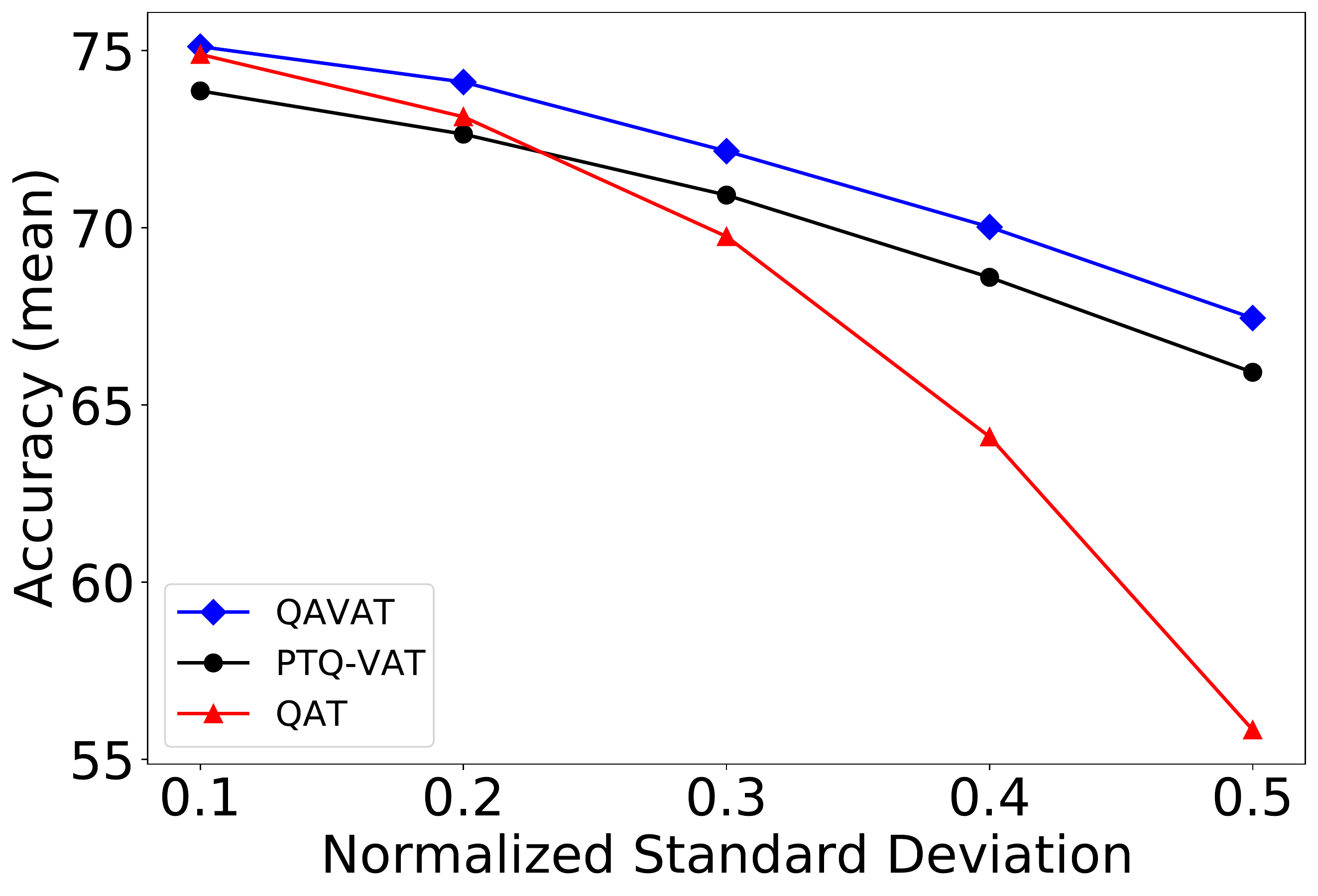}
         \caption{A8W4, weight-proportional}
         \label{fig:resnet_A8W4_mul}
     \end{subfigure}
     \begin{subfigure}[b]{0.24\textwidth}
         \centering
         \includegraphics[width=\textwidth]{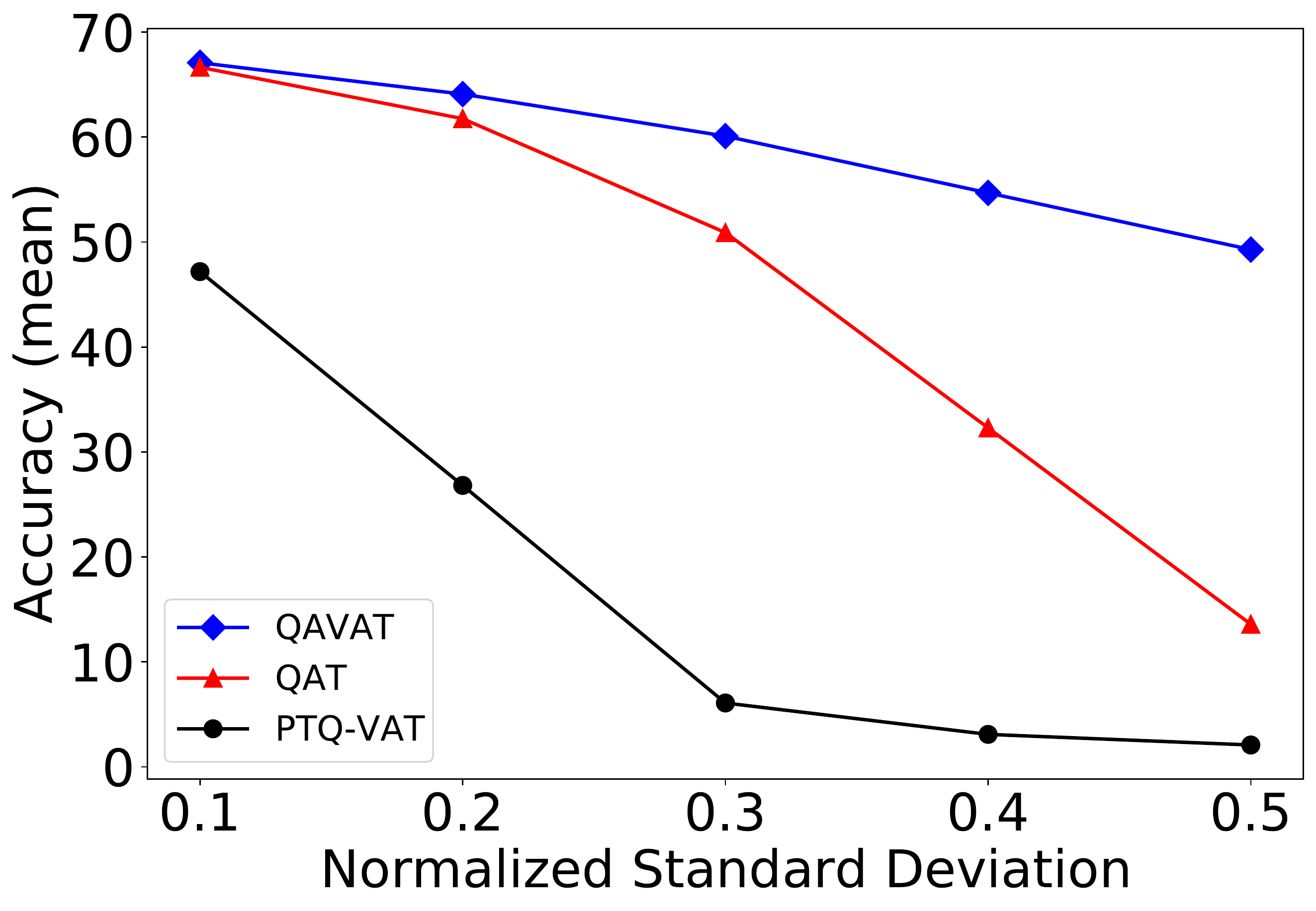}
         \caption{A4W2, layer-fixed}
         \label{fig:resnet_A4W2_add}
     \end{subfigure}
     \hfill
     \begin{subfigure}[b]{0.24\textwidth}
         \centering
         \includegraphics[width=\textwidth]{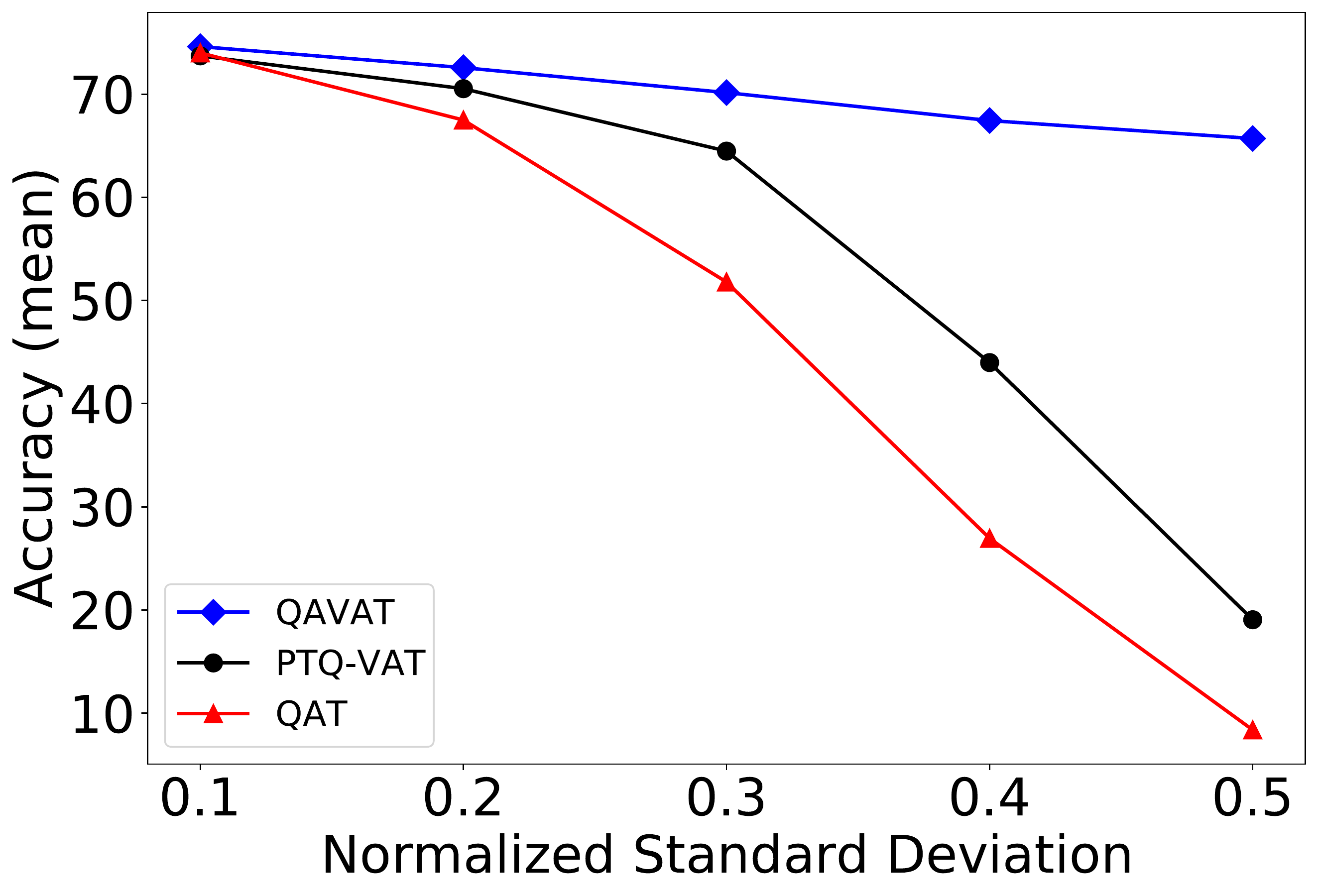}
         \caption{A8W4, layer-fixed}
         \label{fig:resnet_A8W4_add}
     \end{subfigure}
        \caption{QAVAT outperforms PTQ-VAT and QAT.}
        \label{fig:qavat_resnet_fig}
\end{figure}

This set of experiments assumes within-chip variations only.
Table \ref{table:qavat_lowHighVar} presents the results at the lowest and the highest level of variation (for the layer-fixed variance model).
Fig. \ref{fig:qavat_resnet_fig} shows the complete results of ResNet-18/CIFAR-100 experiment.

\emph{Overall, QAVAT is able to produce models that show only modest degradation as the variation increases and for different bitwidths.}
In contrast, QAT models degrade substantially at higher variances. This degradation is more severe in higher bitwidth models (A8W4): the models quantized more aggressively appear to be intrinsically more robust against variations.
For A8W4, the maximum improvement produced by QAVAT is 
$24.9\%$ on  VGG-11 and $11.6\%$ on ResNet-18, under weight-proportional variance. 

\begin{table}[ht]
\begin{tabular}{wl{30pt}wl{12pt}wl{19pt}wl{19pt}wl{19pt}wl{19pt}wl{19pt}wl{19pt}}
\multicolumn{2}{c|}{}&\multicolumn{3}{|c|}{$\sigma=0.1$}&\multicolumn{3}{|c|}{$\sigma=0.5$}\\[0.3ex]
\hline\hline
Model                       & A/W   &VAT    &QAT    &QAVAT  &VAT    &QAT    &QAVAT\\[0.3ex]
\hline\hline
\multirow{2}{*}{ResNet-18}  & 4/2   &47.18  &66.65  &67.08  &2.08   &13.58  &49.28\\[0.3ex]
                            & 8/4   &73.71  &74.00  &74.61  &19.05  &8.37   &65.70\\[0.3ex]
\hline
\multirow{2}{*}{VGG-11 }    & 4/2   &53.76  &87.10  &87.21  &29.72  &68.36  &79.65\\[0.3ex]
                            & 8/4   &88.91  &88.42  &89.00  &77.70  &37.88  &83.09\\[0.3ex]
\hline
\multirow{1}{*}{LeNet-5}    & 2/2   &62.75  &98.21  &98.33  &53.82  &90.03  &96.38\\[0.3ex]
\end{tabular}
\caption{QAVAT at the lowest and the highest level of variability (within-chip variability, layer-fixed variance variance).}
\label{table:qavat_lowHighVar}
\end{table}

Models produced by PTQ-VAT exhibit significant quality degradation at low bitwidths. 
Across all variability levels, the advantage of QAVAT over PTQ-VAT is much larger for A4W2.
For A4W2, on average, across 5 variation levels, QAVAT produces models that are better than PTQ-VAT models by $21.8\%$ on VGG-11 and by $24.9\%$ on ResNet-18,
under weight-proportional variance.

We also study the impact of multi-sampling in QAVAT, Fig. \ref{fig:nv_VGG}. 
Under both A8W4 and A4W2 configurations, mean accuracy of model is improved (by $\sim0.9\%$ at $\sigma=0.3$ and by $\sim1.3\%$ at $\sigma=0.5$).
The gains saturate around 5 samples.

\subsection{Scenario 2: Equal Within-chip and Between-chip Variations}
We next study a more realistic setting in which both between-chip and within-chip components of variation are of equal magnitude: $\sigma_{B}=\sigma_{W}$.
We call this the ``mixed-type'' variation.
In this case, QAVAT is no longer able to produce robust models, as shown in Fig. \ref{fig:withinVsMix}.
\begin{center}
    \centering
    \captionsetup{type=figure}
    \includegraphics[width=.3\textwidth]{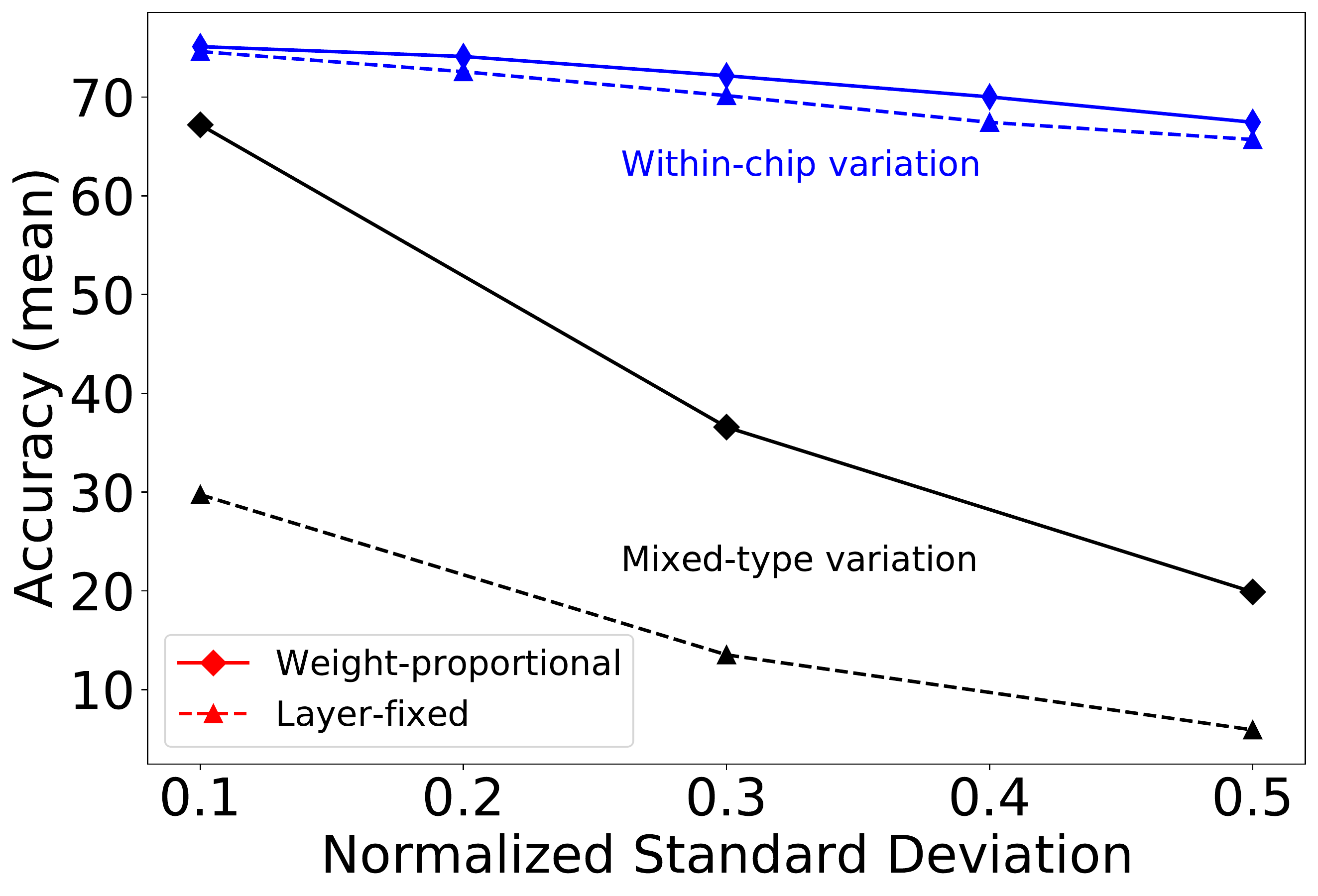}
    \caption{ResNet-18 on CIFAR-100: QAVAT under two scenarios: (1) within-chip variation only, and (2) mixed-type variation. 
    On both variability models (layer-fixed and weight-proportional), degradation in (2) is much more destructive.
    }
	\label{fig:withinVsMix}
\end{center}%

\emph{The self-tuning structure is extremely effective in producing deployable models of much higher accuracy.}
Under $\sigma_{tot}=0.5$ (weight-proportional variance), the best QAVAT-trained models has $36\%$ accuracy loss on VGG-11 and $54\%$ accuracy loss on ResNet-18 compared to a variation-free case.
Under the same setting, self-tuning reduces the accuracy loss on VGG-11 and ResNet-18 to $9.1\%$ and $9.3\%$, respectively.

\begin{figure}[h]
     \centering
     \begin{subfigure}[b]{0.24\textwidth}
         \centering
         \includegraphics[width=\textwidth]{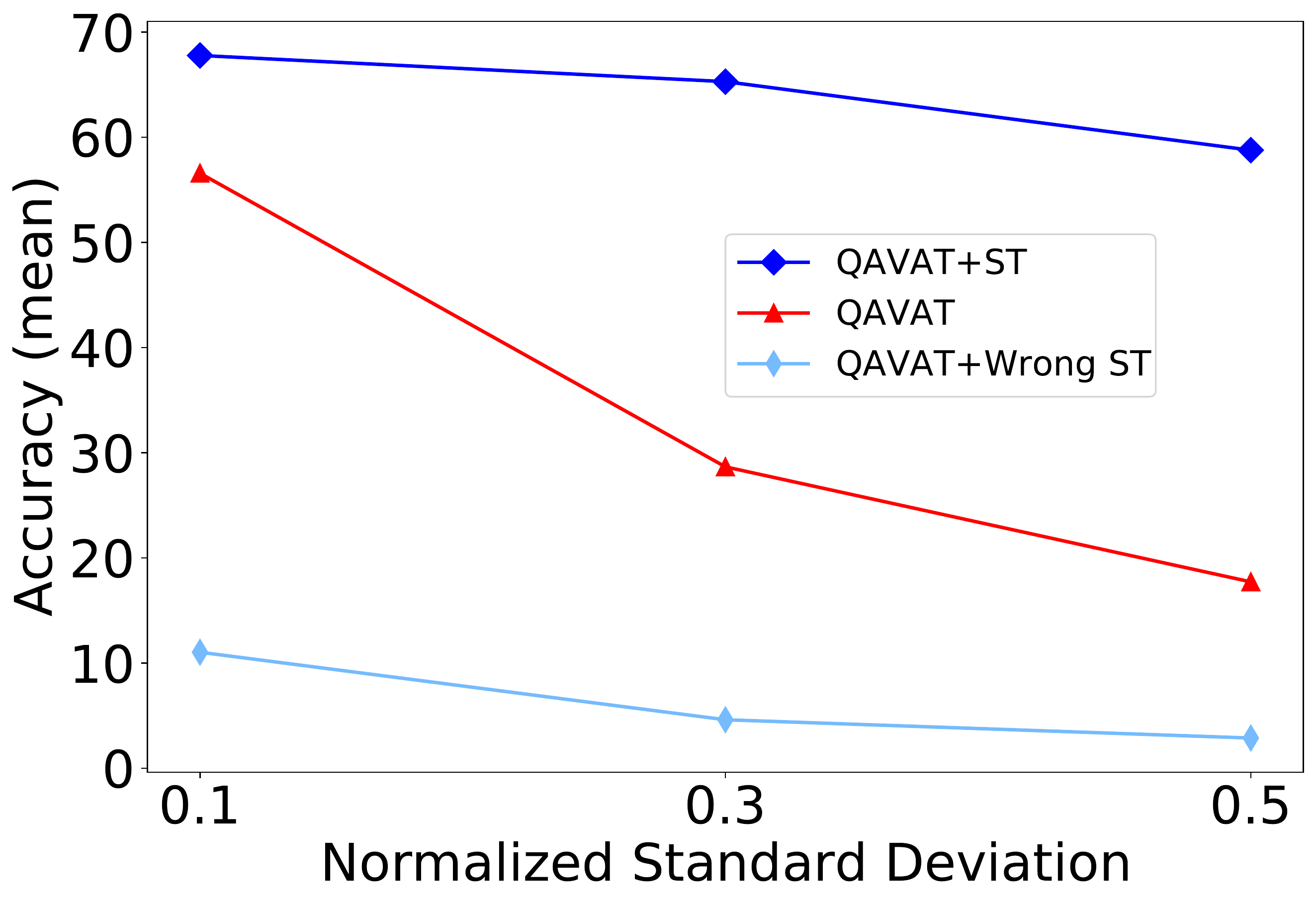}
         \caption{Weight-proportional}
     \end{subfigure}
     \hfill
     \begin{subfigure}[b]{0.24\textwidth}
         \centering
         \includegraphics[width=\textwidth]{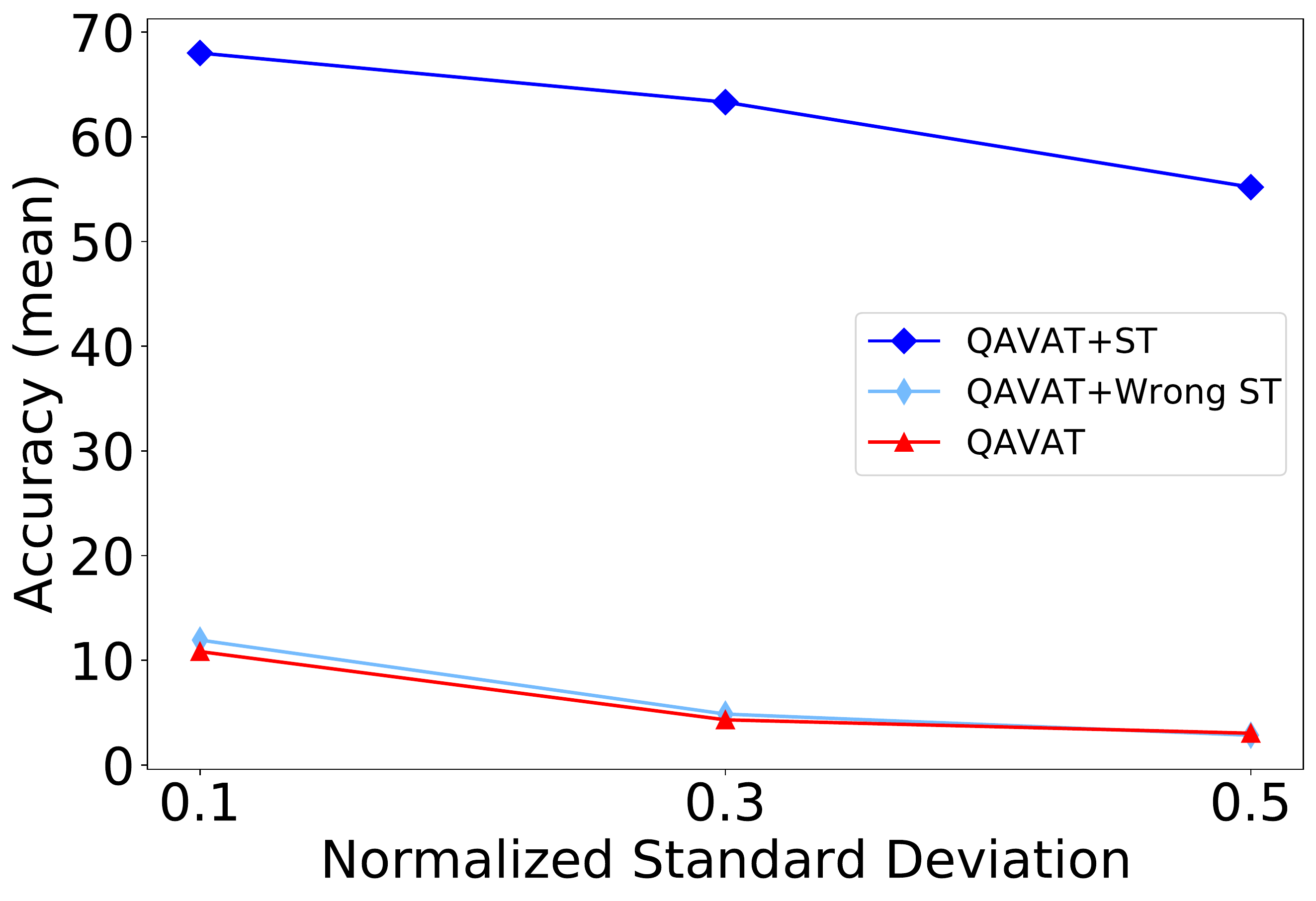}
         \caption{Layer-fixed}
     \end{subfigure}
     
        \caption{Proper application of self-tuning (ST) prevents quality loss: A4W2 ResNet-18, mixed-type variation.}
        \label{fig:calib_resnet_fig}
\end{figure}

Fig. \ref{fig:calib_resnet_fig} shows the performance of a self-tuning architecture for A4W2 ResNet-18.
Table \ref{table:calibTable1} present the results for A8W4 models.
By default, QAVAT+ST uses $10^3$ GTM cells (per model/chip) and 1 LTM column (per layer).
For the layer-fixed variance model at $\sigma_{tot}=0.3,0.5$, QAVAT+ST uses $10^5$ GTM cells and 16 LTM columns to achieve the results shown. 
As presented in Fig. \ref{fig:selfCalib}, the layer-fixed variance requires the GTM+LTM configuration of ST, while the weight-proportional variance requires only a GTM.
We show that applying the wrong type of self-tuning, i.e., using a GTM-only ST for layer-fixed variance, and vice-versa, is not helpful and leads to drastic degradation (``QAVAT+Wrong ST'' in the figure and table).

The computation overhead of self-tuning is the ratio of inference-time FLOPs in all tuning modules (GTM+LTMs) compared to the base model.
On ResNet-18, assuming $10^5$ GTM cells are used, the overhead is $\sim0.3\%$ when LTM=1, $\sim 2.2\%$ when LTM=8, and $\sim4.4\%$ when LTM=16. 
\begin{table}[ht]
\begin{tabular}{wl{60pt}wc{18pt}wc{18pt}wc{18pt}wc{18pt}wc{18pt}wc{18pt}}
\multicolumn{1}{c|}{}&\multicolumn{3}{|c|}{VGG-11}&\multicolumn{3}{|c|}{ResNet-18}\\[0.3ex]
\hline\hline
$\sigma_{tot}$  & 0.1  & 0.3 & 0.5 & 0.1  & 0.3 & 0.5\\[0.3ex]
\hline\hline
QAVAT           &88.59  &70.75  &54.70  &67.19  &36.58  &19.89\\[0.3ex]
\hline
QAVAT+ST        &90.05  &88.09  &81.90  &75.35  &73.39  &66.58\\[0.3ex]
\hline
QAVAT+Wrong ST  &44.70  &23.06  &17.33  &14.32  &5.26   &3.78 \\[0.3ex]
\end{tabular}
\caption{Proper application of self-tuning (ST) prevents quality loss: mixed-type variation, weight-proportional variance.}
\label{table:calibTable1}
\end{table}

\begin{figure}[h]
    
     \centering
     \begin{subfigure}[b]{0.24\textwidth}
         \includegraphics[width=\textwidth]{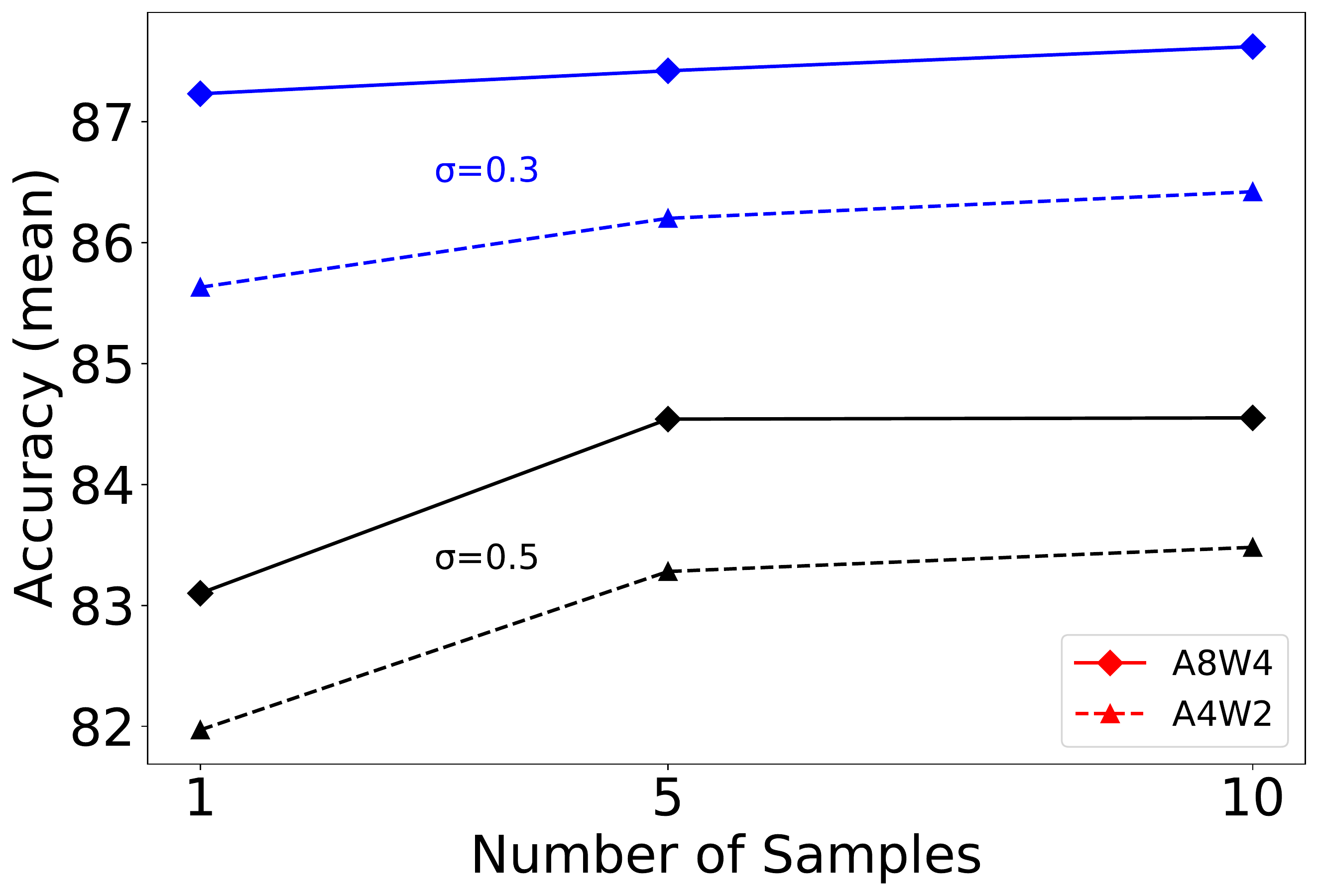}
         \caption{Impact of multi-sampling (VGG-11, within-chip variation).}
         \label{fig:nv_VGG}
    \end{subfigure}
     \hfill
     \begin{subfigure}[b]{0.24\textwidth}
         \centering
            \includegraphics[width=\textwidth]{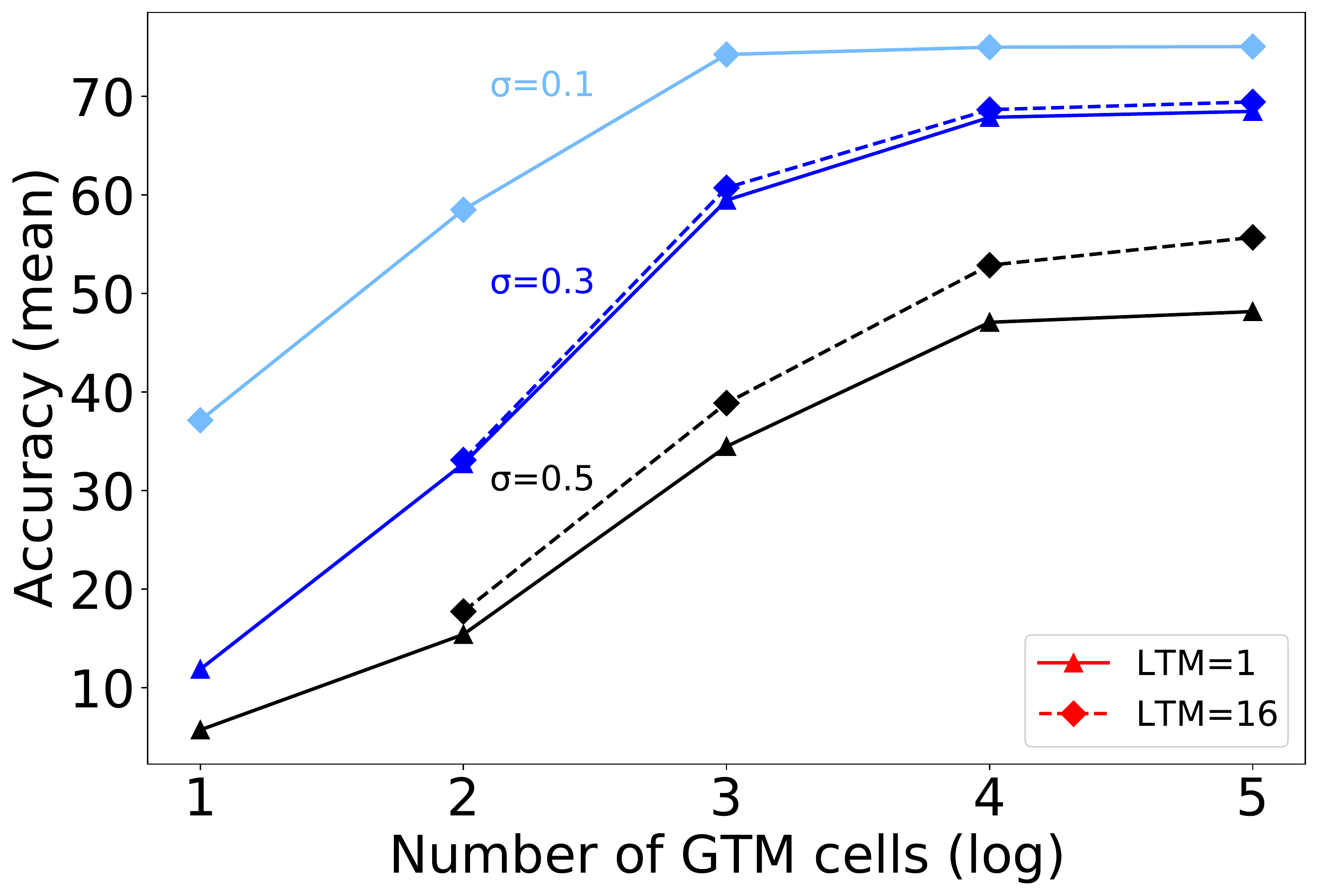}
            \caption{Impact of ST size (ResNet-18, mixed-type variation).}
            \label{fig:n_calib_resnet}
     \end{subfigure}
        \caption{Impact of QAVAT and ST parameters on model quality.}
        \label{fig:n_calib}
\end{figure}

Performance of self-tuning improves with increasing number of cells in GTM, Fig. \ref{fig:n_calib_resnet} (layer-fixed variance).
However, there is a diminishing return in model improvement.
Larger variance requires a higher number of GTM cells before improvements diminish.
Similar trade-off works for variability with weight-proportional variance.
For self-tuning of variability with layer-fixed variance, increasing the number of LTM columns is an orthogonal way of reducing estimation variance and improving results.
As shown in the figure, the usefulness of more LTMs is clear under the highest variance level ($\sigma=0.5$).

\section{Conclusion}
We present first a general variability- and quantization-aware training algorithm (QAVAT) to mitigate the degradation of DNN performance under analog PIM non-idealities.
QAVAT is more effective at reducing the degradation than alternatives on three computer vision networks and datasets and a wide range of variability and quantization levels.
We also demonstrate a novel lightweight self-tuning architecture that can be used on arbitrary DNNs.
It dynamically adjusts layers' outputs to dramatically reduce degradation under correlated variations.
\newline
\bibliographystyle{unsrt}  
\bibliography{references}

\end{document}